\documentclass{article}

\PassOptionsToPackage{numbers,compress}{natbib}

\usepackage[preprint]{submission}

\usepackage[utf8]{inputenc} 
\usepackage[T1]{fontenc}    
\usepackage{hyperref}       
\usepackage{url}            
\usepackage{booktabs}       
\usepackage{amsfonts}       
\usepackage{float} 
\usepackage{nicefrac}       
\usepackage{microtype}      
\usepackage{xcolor}         
\usepackage{bm}
\usepackage{graphicx}
\usepackage{amsmath}
\usepackage{amssymb}
\usepackage{mathtools}
\usepackage{amsthm}
\usepackage{multirow}
\usepackage{hyperref}
\usepackage{algorithm}
\usepackage{algorithmicx}
\usepackage{algpseudocode}
\usepackage{adjustbox}
\usepackage{wrapfig}
\usepackage{pifont}

\title{DopQ-ViT: Towards Distribution-Friendly and Outlier-Aware Post-Training Quantization \\
for Vision Transformers}

\author{%
  Lianwei Yang\footnotemark[1] $^{\ 1,2}$, Haisong Gong\footnotemark[1] $^{\ 1,2}$, Haokun Lin\footnotemark[1] $^{\ 1,2,4}$, \\ \textbf{Yichen Wu $^{3}$, Caifeng Shan$^{\ 5}$, Zhenan Sun$^{\ 1,2}$, Qingyi Gu\footnotemark[2] $^{\ 1}$ }\\
  $^{1}$ Institute of Automation, Chinese Academy of Sciences \\ 
  $^{2}$ School of Artificial Intelligence, University of Chinese Academy of Sciences \\
  $^{3}$ Center for Advanced Medical Computing and Analysis, Harvard University \\
  $^{4}$ Department of Computer Science, City University of Hong Kong \\
  $^{5}$ School of Intelligence Science and Technology, Nanjing University \\
}

\begin{document}
\renewcommand{\thefootnote}{\fnsymbol{footnote}}
\footnotetext[1]{Equal contribution.}  
\footnotetext[2]{Corresponding authors.}
\vspace{-0.6cm}
\maketitle

\begin{abstract}
Vision Transformers (ViTs) have gained significant attention, but their high computing cost limits the practical applications.
While post-training quantization (PTQ) reduces model size and speeds up inference, it often degrades performance, especially in low-bit settings.
We identify two key reasons for the performance degradation: \textbf{1)} existing quantization methods fail to align with the power-law distribution of post-Softmax activations, and \textbf{2)} reparameterizing post-LayerNorm activations leads to a performance drop due to the significant influence of outliers in the scaling factors.
To address these challenges, we propose \textbf{DopQ-ViT}, a \textbf{D}istribution-friendly and \textbf{O}utlier-aware \textbf{P}ost-training \textbf{Q}uantization method for ViTs.
First, DopQ-ViT introduces the Tan Quantizer (TanQ), which better preserves the power-law distribution of post-Softmax activations by focusing more on values near 1.
Second, DopQ-ViT presents the MAD-guided Optimal Scaling Factor (MOSF), which selects the optimal scaling factor without introducing additional calculations.
Extensive experiments across various ViT models and quantization settings demonstrate that DopQ-ViT, with the help of TanQ and MOSF, outperforms previous PTQ methods on both classification and detection tasks.
\end{abstract}

\section{Introduction}
\label{sec:Introduction}
Vision Transformers (ViTs) have demonstrated strong capabilities in image classification~\cite{bhojanapalli2021understanding}, object detection~\cite{li2022exploring}, and instance segmentation~\cite{wang2021end}.
However, the performance of ViTs is highly dependent on computational resources and memory, which poses challenges for deploying them on edge devices.
This has led to various efforts to compress ViTs through techniques like pruning~\cite{zhu2021vision, lin2024mope}, distillation~\cite{wu2022tinyvit}, and quantization~\cite{liu2021post}.
Among these, network quantization, which converts network weights or activations from floating-point to fixed-point formats, has become a mainstream method for reducing model size and computational requirements.
To maintain the performance of quantized models, several approaches adopt Quantization-Aware Training (QAT)~\cite{nagel2022overcoming}, which involves end-to-end retraining to recover performance.
However, QAT relies on the training set, resulting in high memory usage and long training time, making it difficult to implement in practical applications.
In contrast, Post-Training Quantization (PTQ)~\cite{lin2022fqvit, Repq-vit, zhongyunshan} offers a promising alternative by utilizing only a small subset of training data to minimize the reconstruction error introduced during quantization.

\begin{figure*}[t]
\begin{center}
\centerline{\includegraphics[width=1.0\textwidth]{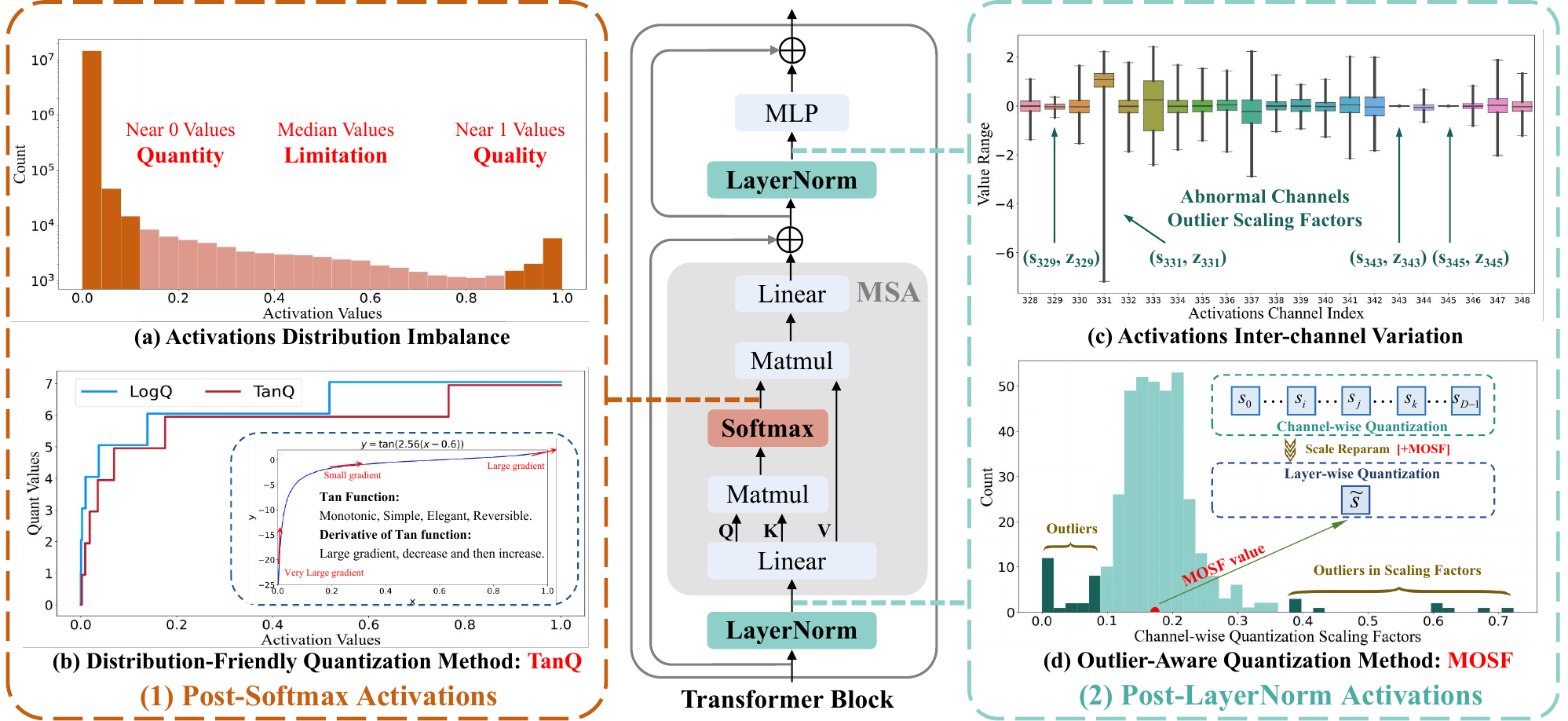}}
\vskip -2mm
\caption{Overview of DopQ-ViT. (1) DopQ-ViT proposes the distribution-friendly Tan Quantizer to address the imbalance in the distribution of post-Softmax activations, thereby enhancing the practicality of low-bit quantization. (2) DopQ-ViT introduces an outlier-aware method MOSF to address the inter-channel variation in post-LayerNorm activations, aiming to select the MAD-guided optimal scaling factor and mitigate accuracy degradation during the reparameterization process.}
\label{fig:overview}
\end{center}
\vskip -10mm
\end{figure*}
The main challenge of post-training quantization for ViTs is the significant performance drop under low-bit weight-activation quantization.
Previous works~\cite{lin2022fqvit, Repq-vit} have identified two key reasons for this issue. 
First, the imbalanced activations after post-Softmax make the uniform quantizer ineffective. 
To address this, PTQ4ViT~\cite{yuan2022ptq4vit} introduces the twin uniform quantizer, while FQ-ViT~\cite{lin2022fqvit} proposes the Log2 quantizer. 
Additionally, APQ-ViT~\cite{ding2022towards} introduces Matthew-effect Preserving Quantization to better align with the data distribution, and I\&S-ViT~\cite{zhongyunshan} uses the Shift-Uniform-Log2 Quantizer (SULQ) to enhance quantization efficiency.
Second, post-LayerNorm activations are better suited for per-channel quantization due to channel-wise variations, as shown in Figure~\ref{fig:overview}(c). 
However, to speed up inference, most current methods use layer-wise quantization for all activations. 
Thus, RepQ-ViT~\cite{Repq-vit} proposes reparameterizing the channel-wise scaling factors and zero points into a single value by taking the mean of these values.
While these techniques significantly enhance the capacity of quantized models, they still suffer from degradation in performance under 3-bit settings.

We conduct detailed preliminary experiments and identify two main issues contributing to the degradation: 
\textbf{1)}. We observe that a small set of values close to 1 in the post-Softmax activations are important as removing them destroys the quantized models. 
Previous Log-based quantizers focus primarily on values near 0 and fail to effectively represent these important values near 1. 
\textbf{2)}. We discover that some channels in post-LayerNorm activations exhibit abnormal value ranges, leading to outliers in the channel-wise scaling factors and zero points. 
These outliers hinder the effectiveness of reparameterization and contribute to the poor performance of RepQ-ViT under 3-bit quantization.

To address these issues, we propose a novel and efficient PTQ method for ViTs, named \textbf{DopQ-ViT}.
First, we introduce a distribution-friendly \textbf{Tan Quantizer (TanQ)} that effectively preserves both values near 0 and values near 1 in the post-Softmax activations. TanQ focuses more on values near 1, accurately fitting the distribution, and ensures better representation across the entire range of activations.
Second, we observe that directly using the mean value to transition post-LayerNorm activations from channel-wise to layer-wise quantization is ineffective. 
To deal with abnormal values, we propose a training-free search method called \textbf{MAD-guided Optimal Scaling Factor (MOSF)} to identify the optimal scaling factor.
MOSF leverages the Mean Absolute Deviation (MAD) as an evaluation metric to determine the scaling factor, without introducing additional overhead.
This scaling factor enables precise quantization for most channels, while compensating for errors caused by abnormal channels.
With the help of MOSF, DopQ-ViT reparameterizes the post-LayerNorm activations more accurately, significantly improving the performance of low-bit quantized models.

Our main contributions are summarized as follows.
\textbf{1)}. We identify two key issues with current PTQ methods for ViTs: (i) existing quantizers do not align well with the distribution of post-Softmax activations, leading to poor efficiency and (ii) the effectiveness of reparameterization is affected by abnormal channels in post-LayerNorm activations and outliers in scaling factors.
    %
\textbf{2)}. We propose DopQ-ViT with two novel methods: (i) Tan Quantizer (TanQ), which places greater focus on values near 1 to better match the distribution of post-Softmax activations, and (ii) MAD-guided Optimal Scaling Factor (MOSF), which selects a more suitable scaling factor for reparameterizing post-LayerNorm activations.
\textbf{3)}. We evaluate DopQ-ViT on image classification and object detection tasks across various models and bit settings. Compared with existing PTQ methods, DopQ-ViT exhibits significant advantages in accuracy and speed, with only a slight additional computational overhead.

\section{Preliminaries}
\label{sec:Preliminaries}
\subsection{Vision Transformers} 
\label{sec:Structure of ViTs}
\textbf{Structure of ViTs.} ViTs are formed by stacking multiple transformer encoder blocks, 
with each encoder consisting of a multi-head self-attention (MSA) and a multi-layer perceptron (MLP).
%
For an input image, ViTs divide it into $N$ patches.
Then, the embedding layer converts these patches into a sequence of $D$-dimensional vectors.
This vector is fed into $L$ blocks to complete the subsequent operations.
Specifically, for the $l_{th}$ block, when the input is $\bm{X}_{l}$, the process is executed as follows:
\begin{equation}
\begin{aligned}
\bm{Y}_{l} &=  \bm{X}_{l} + \text{MSA}(\text{LayerNorm}(\bm{X}_{l})) . \\
\bm{X}_{l+1} &= \bm{Y}_{l} + \text{MLP}(\text{LayerNorm}(\bm{Y}_{l})) .
\end{aligned}
\end{equation}
\textbf{MSA.} The MSA consists of $h$ self-attention heads that capture correlations between patches.
MSA obtains the query ($\bm{Q}$), key ($\bm{K}$), and value ($\bm{V}$) vectors through linear mappings.
Then, applying Softmax to the query and key yields the attention scores. The final output of MSA is obtained by concatenating the outputs from all heads.
For the $i_{th}$ head, when the input is $\bm{X}^{'}$, the process is:
\begin{equation}
\begin{aligned}
[\bm{Q}_{i}, \bm{K}_{i}, \bm{V}_{i} ] &= \bm{X}^{'}\bm{W}^{QKV}+\bm{b}^{QKV} .\\
\text{Attn}_{i} &= \text{Softmax}\left({\bm{Q}_{i}\cdot \bm{K}_{i}^T}/{\sqrt{D_{h}}}\right)\bm{V}_{i} .\\
\text{MSA}(\bm{X}^{'}) &= [\ \text{Attn}_{1}, \text{Attn}_{2}, ...,\text{Attn}_{h}]\ \bm{W}^O +\bm{b}^O   .
\end{aligned}
\end{equation}
\textbf{MLP.} The MLP learns feature representations by projecting them into a high-dimensional space via fully connected layers and GELU operations. 
$\bm{W}^{1}$, $\bm{W}^{2}$, $\bm{b}^{1}$, and $\bm{b}^{2}$ represent the weights and biases of the two fully connected layers in the MLP.
When the input is $\bm{Y}^{'}$, the process is:
\begin{align}
\text{MLP}(\bm{Y}^{'}) = \text{GELU}\left(\bm{Y}^{'}\bm{W}^{1} + \bm{b}^{1} \right)\bm{W}^{2} + \bm{b}^{2}.
\end{align}

\subsection{Existing Quantizers for post-Softmax Activations} 
\label{Existing quantizers for post-Softmax activations}
The Uniform Quantizer (UQ) maps floating-point data to integer values and is widely used. 
Here, $s$ denotes the scaling factor, $z$ represents the zero point, and the clamp function is used to limit the range.
Given the bit-width $b$, the processes of quant and dequant are defined as follows:
\begin{equation}
\begin{aligned}
\label{eq: Initialize model1}
\bm{x}_{q} = \text{clamp}\left(\left\lfloor\frac{\bm{x}}{s}\right\rceil + z, 0, 2^{b}-1\right) ,& \quad {\bm{x}_{f}} = s \cdot (\bm{x}_{q} - z) \approx \bm{x} .\\
\text{where}\quad{s} = \frac{\max(\bm{x}) - \min(\bm{x})}{2^b-1},& \quad  {z} = \left\lfloor -\frac{\min(\bm{x})}{s} \right\rceil.
\end{aligned}
\end{equation}

However, when processing imbalanced activations, uniform quantizer introduces significant quantization errors under low-bit settings.
In ViTs, the attention probabilities computed by Softmax, denoted as post-Softmax activations, fall in the range $[0,1]$. 
Their distribution is neither Gaussian nor Laplacian, but follows a long-tail or power-law distribution.
Such extreme distributions present substantial challenges for PTQ.
FQ-ViT~\cite{lin2022fqvit} introduces an innovative Log2 quantizer that allocates more quantization intervals to the dense regions near 0 in the activations.
RepQ-ViT~\cite{Repq-vit} further enhances performance by proposing the Log$\sqrt{2}$ quantizer.
I\&S-ViT~\cite{zhongyunshan} presents the Shift-Uniform-Log2 Quantizer (SULQ), which introduces a shift bias $\eta$ to optimize the efficiency of the Log2 quantizer.
\begin{equation}
\begin{aligned}
\label{eq: Initialize model3}
\text{Log$_aQ(\bm{x}, b)$: } \bm{x}_{q} &= \text{clamp}\left(\left\lfloor -\log_a\frac{\bm{x}}{s} \right\rceil , 0, 2^{b}-1\right), a \in \{2, \sqrt2\} .\\
\text{SULQ$(\bm{x}, b)$: }  \bm{x}_{q} &= \text{UQ}(-\log_2(\bm{x}+\eta), b) .
\end{aligned}
\end{equation}
\subsection{Scale Reparameterization for Post-LayerNorm Activations}
\label{Scale Reparam for post-LayerNorm Activations}
Following RepQ-ViT, we apply channel-wise quantization for weights and layer-wise quantization for activations.
We present the visualization of post-LayerNorm activations in Figure~\ref{fig:overview}(c), where we observe significant inter-channel variations.
Thus, layer-wise quantization suffers from considerable accuracy degradation because a single scaling factor is unable to effectively capture the inter-channel variations.
In contrast, channel-wise quantization better accommodates these activations but requires dedicated hardware support and introduces additional computational overhead.
To strike a balance between the efficiency of layer-wise quantization and the accuracy of channel-wise quantization, RepQ-ViT introduces the scale reparameterization.
It first applies channel-wise quantization to obtain scaling factors $\bm{s}\in \mathbb{R}^{D}$ and zero points $\bm{z}\in \mathbb{R}^{D}$. 
Then calculate the values of $\tilde{s}=\text{Mean}(\bm{s}) \in \mathbb{R}^{1}$ and $\tilde{z}=\text{Mean}(\bm{z})\in \mathbb{R}^{1}$ for layer-wise quantization.
The reparameterization is achieved by adjusting the affine parameters $\bm{\beta}$ and $\bm{\gamma}$ of LayerNorm, where $LN(\bm{X}) = \frac{\bm{X} - E(\bm{X})}{Var(\bm{X})} * \bm{\gamma} + \bm{\beta}$, to new vectors $\widetilde{\bm{\beta}}$ and $\widetilde{\bm{\gamma}}$.
Specifically, the variation factors are computed using
$\bm{r}_1=\bm{s}/\tilde{\bm{s}}$ and $\bm{r}_2=\bm{z}-\tilde{\bm{z}}$, where $\bm{\tilde{s}} = \tilde{s} \cdot \bm{1}$ and $\bm{\tilde{z}} = \tilde{z} \cdot \bm{1}$.
Here, $\odot$ denotes Hadamard product. $\widetilde{\bm{\beta}}$ and $\widetilde{\bm{\gamma}}$ can be derived through
\begin{align}
  \label{eq:affine}
  \widetilde{\bm{\beta}} &= \frac{\bm{\beta}+\bm{s}\odot \bm{r}_2}{\bm{r}_1},  \quad 
  \widetilde{\bm{\gamma}} = \frac{\bm{\gamma}}{\bm{r}_1}  .
\end{align}
To address the distribution shift caused by the above operations, RepQ-ViT adjusts the weights of the subsequent layer by $\widetilde{\bm{W}}_{[:,j]} = \bm{r}_1\odot\bm{W}_{[:,j]}$ and $\widetilde{\bm{b}}_j = \bm{b}_j - (\bm{s}\odot \bm{r}_2) \bm{W}_{[:,j]}$.
This technique reduces the channel-wise discrepancies, leading to great performance with 6-bit weight-activation quantization. However, existing baselines still face notable performance drops when quantized to low-bit settings.

\section{Motivation}
\label{sec:Motivation}
\subsection{Distribution Imbalance in Post-Softmax Activations}
\label{Post-Softmax Activations Distribution Imbalance}
\textbf{OB 1.} 
Post-Softmax activations follow an unbalanced power-law distribution, and existing quantizers cannot simultaneously focus on values close to 0 and 1, resulting in performance degradation.

\begin{wraptable}[8]{r}{7.5cm}
\vspace{-6mm}
\centering
\small
\caption{The ImageNet Top-1 accuracy(\%) after clipping 1\% of post-Softmax activations.}
\begin{adjustbox}{width=\linewidth}
\begin{tabular}{cc}
\toprule
\multirow{2.5}{*}{\textbf{Clipping Position}}  & \multicolumn{1}{c}{\textbf{Model quantized with W4A4}}\\
& ViT-S  ~~~~ ViT-B ~~~~ DeiT-T ~~~ DeiT-S  \\
\midrule
Full-Precision~~~~ &81.39~~~~~~84.54 ~~~~~72.21~~~~~~ 79.85\\
Clipping 1\% Near 0 &62.99 ~~~~ 70.73 ~~~~~ 57.19 ~~~~~  68.99 \\
Clipping 1\% Near 1 &25.09 ~~~~~ 2.20 ~~~~~~ 43.05 ~~~~~ 45.89  \\ 
\bottomrule
\end{tabular}
\label{tab:clip_near}
\end{adjustbox}
\end{wraptable}
Figure~\ref{fig:overview}(a) presents the histogram of post-Softmax activations. 
It is evident that the majority of activation values are concentrated near 0, with only a small fraction approaching 1. 
Given the large concentration of activation values close to 0, it is crucial to allocate more quantization intervals to this region to improve differentiation~\citep{lin2022fqvit,Repq-vit}.
%
We conducted preliminary experiments to explore the role of activations near 1. 
In Table~\ref{tab:clip_near}, we exclude 1\% of the activations both close to 0 and close to 1 for post-Softmax activations, respectively, and then evaluate the impact of these activations in quantization.
%
We find that quantized models with activations near 1 removed perform significantly worse than those with activations near 0 excluded. 
Specifically, for the ViT-B model, the accuracy drops to just 2.20\%. 
This suggests that activations near 1 play a crucial role, and we must consider how to preserve these values during quantization.

These existing Log-based quantizers~\citep{lin2022fqvit,Repq-vit} allocate more quantization intervals to regions near 0, resulting in better performance compared to the uniform quantizer. 
However, they neglect the values near 1, leading to poor performance under low-bit settings.
SULQ tries to address this by introducing a shift bias $\eta$, but it still suffers from inherent limitations as a Log-based quantizer. 
Theoretically, Softmax compresses small values and amplifies large ones, creating a "Matthew Effect" that helps stabilize activations. 
Existing Log-based quantizers overlook this effect, allocating insufficient quantization intervals to values near 1, which effectively destroys information in the larger values.
Therefore, a more suitable quantizer is needed that can simultaneously address both values near 0 and those near 1.
\subsection{Channel Variation in Post-LayerNorm Activations}
\label{Post-LayerNorm Activations Channel Variation}
\textbf{OB 2.} After scale reparameterization of the post-LayerNorm activations, the model is negatively affected by abnormal channels, leading to a decrease in accuracy, especially under low-bit settings.

Post-LayerNorm activations are sensitive to low-bit quantization. 
We further analyze the scale reparameterization under 3-bit activation quantization.
When quantizing these activations with channel-wise quantization, the accuracy of ViT-S on ImageNet is 55.70\%. When directly applying RepQ-ViT's layer-wise reparameterization, the accuracy drops significantly to 40.88\%. This 14.82\% drop in accuracy highlights the limitation of such reparameterization in preserving model performance.

\begin{wraptable}[9]{r}{7.5cm}
\vspace{-7mm}
\centering
\small
\caption{The ImageNet Top-1 accuracy(\%) of the ViT-S after 3-bit activation quantization, clipping a\% smallest and b\% largest scaling factors.}
\begin{adjustbox}{width=\linewidth}
\begin{tabular}{cc}
\toprule
\multirow{2.5}{*}{\textbf{Quantization}}  & \multirow{1}{*}{\textbf{Clipping Ratio ~~ (a\%, b\%)}}\\
& (0,0) ~~~~ (5,15) ~~~ (10,10) ~~~ (15,5)  \\
\midrule
Channel-wise    &  55.70~~~~ 55.70~~~~~ 55.70~~~~~ 55.70  \\ 
\multirow{1}*{$\downarrow$Reparam} & 14.82$\downarrow$ ~~~13.23$\downarrow$ ~~~12.45$\downarrow$ ~~~~9.22$\downarrow$ \\
Layer-wise    &40.88 ~~~ 42.47 ~~~~ 43.25 ~~~~  46.48\\
\bottomrule
\end{tabular}
\label{tab:table2}
\end{adjustbox}
\end{wraptable}
Since RepQ-ViT is based on rigorous mathematical derivation, we analyze that the drop in accuracy arises from the method of determining the scaling factor for layer-wise quantization.
From Figure~\ref{fig:overview}(c), we observe that some channels exhibit abnormal value ranges, resulting in scaling factors that are either too large or too small. 
These abnormal values in scaling factors, denoted as outliers, affect the mean, making $\tilde{s} = \text{Mean}(\mathbf{s})$ less representative of the entire set.
We remove the minimum $a\%$ and the maximum $b\%$ of scaling factors during the calculation process, and then reparameterize by taking the mean of the remaining values. 
As shown in Table~\ref{tab:table2}, this operation slows the rate of accuracy degradation.
These results indicate that abnormal outliers influence the reparameterization, and ignoring these outliers suppresses the decrease in accuracy.
The scaling factor $\tilde{s} = \text{Mean}(\mathbf{s})$ can not effectively represent all channels during layer-wise quantization, leading to an inevitable decrease.
Therefore, selecting the optimal scaling factor for layer-wise quantization is crucial to overcoming the influence of abnormal channels. 
An ideal scaling factor should provide precise quantization for the majority of channels, compensate for the few abnormal channels, and improve model accuracy.
\section{Methodology}
\label{sec:Methodolog}
\subsection{Distribution-Frinendly Quantization for Post-Softmax Activations}
\label{Distribution-Frinendly Quantization for Softmax}
\textbf{Tan Quantizer.} 
Inspired by the above analysis in Section~\ref{Post-Softmax Activations Distribution Imbalance}, we propose the Tan Quantizer (TanQ) to address this issue. As shown in Figure\,\ref{fig:overview}(b), the ideal quantization function has a very large gradient when $x$ is near 0, with the function value approaching negative infinity. 
As $x$ increases, the gradient of the function decreases, and as $x$ nears 1, the gradient increases.
The function is monotonic, simple, and has an inverse function. Fortunately, $\textit{tan}(a(\bm{x}-b))$ perfectly meets these requirements.
The Tan quantizer is simple and efficient, and the quant and dequant processes of TanQ are as follows:
\begin{equation}
\begin{aligned}
\label{eq: Initialize model4}
& \bm{x}_{q} = \text{clamp}\left(\left\lfloor {\frac{{\textit{tan}(a(\bm{x}-b))}}{s}} \right\rceil+z, 0, 2^{b}-1\right) .\\
& {\bm{x}_{f}} = \textit{arctan}(\left\lfloor s \cdot ( \bm{x}_{q}- z)  \right\rceil)/a + b\approx \bm{x} .
\end{aligned}
\end{equation}
TanQ introduces hyperparameters $a$ and $b$, reducing emphasis on intermediate values and focusing more on values close to 0 and near 1.
\textbf{(i)} 
Parameter $a$ adjusts the curvature of the tan function, thereby influencing the number of quantization intervals allocated to values close to 0.
For example, when $a$ is larger, TanQ will allocate more quantization intervals to the region close to 0 to increase discrimination.
\textbf{(ii)} 
Parameter $b$ is responsible for adjusting the focus on values close to 1.
For instance, when $b$ is 1, TanQ does not pay extra attention to values close to 1; when $b$ is 0.5, TanQ focuses equally on values close to 0 and values close to 1.
Therefore, $a$ and $b$ can cooperate with each other to achieve the best quantization effect.
Besides, parameters $a$ and $b$ both need to satisfy constraints, which can be determined through a simple grid search method.
Thus, TanQ enhances quantization efficiency with just a slight additional computational overhead.
The optimization problem is:
\begin{equation}
\label{eq: Initialize model5}
\begin{aligned}
 min\|\bm{x}_{f} -& \bm{x}\|_2 \\
 s.t. \quad b + \pi/2a &> 1  \\
 b - \pi /2a &< 0   \\
 a > 0, b > &0, b < 1  
\end{aligned}
\end{equation}

\textbf{Superiority.} 
We present a visual comparison of the Log quantizer and Tan quantizer in Figure\,\ref{fig:quantization error}.
The Tan quantizer exhibits lower quantization errors. 
TanQ allocates more quantization intervals to values close to 0 and gives greater emphasis to values close to 1, effectively addressing the issue of low quantization efficiency found in Log2 and Log$\sqrt{2}$ quantizers.
TanQ aligns well with the power-law distribution of post-Softmax activations and performs well.
%
%
Besides, TanQ can adjust its parameters $a$ and $b$ to achieve customized quantization for post-Softmax activations of different blocks, thereby improving the performance of the quantized models.
Previous uniform and Log quantizers perform the same quantization for all the post-Softmax activations in ViTs, which limits the performance.
%
%

\begin{wraptable}[5]{r}{6.5cm}
\vspace{-7mm}
\centering
\small
\caption{Hardware efficiency comparison.}
\begin{adjustbox}{width=\linewidth}
\begin{tabular}{l|ccc}
\toprule
\textbf{Method} & \textbf{Latency(ms)} & \textbf{SpeedUp} & \textbf{Error}   \\
\midrule
LogQ 	&230.3 	&1.0x 	&0.07 \\
TanQ 	&14.2 	&16.2x 	&7.6e-16  \\
\bottomrule
\end{tabular}
\label{tab:cordic}
\end{adjustbox}
\end{wraptable}

\textbf{Hardware Implementation.} CORDIC~\cite{chen2020cordic, deng2024efficient, wu2023decision} is an iterative algorithm that calculates the tangent value using addition, subtraction, and bit-shift. 
%
%
To evaluate the hardware performance, we use CORDIC to simulate tan and log calculations on an Intel i7-10510U CPU. We benchmark 1000 data points with 24 CORDIC iterations.
As shown in Table \ref{tab:cordic}, TanQ is significantly faster than the Log quantizer with a much smaller error,
demonstrating the efficiency and hardware friendliness.
%
In CORDIC, we cannot directly use the $log_2(\bm{x})$ but instead employ the $log_2(\bm{x})=ln(\bm{x})/ln(2)$. $ln(\bm{x})$ is computed approximately by $2 \times \textit{arctanh} ( \frac{x-1}{x+1})$. However, the domain of $\textit{arctanh}$ is defined in the range $[-0.8069,0.8069]$~\cite{fang2023scalable,xie2023novel}, limiting the $\bm{x} \in [0.1069, 9.3573]$.
Since the activations range is (0,1), LogQ has difficulty handling activations near 0, resulting in larger errors.
And it requires more calculations.

\subsection{Outlier-Aware Quantization for Post-LayerNorm Activations}
\label{Outlier-Aware Quantization for LayerNorm}
\textbf{MOSF.} Based on the analysis in Section~\ref{Post-LayerNorm Activations Channel Variation}, we propose a method to select the \textbf{M}AD-guided \textbf{O}ptimal \textbf{S}caling \textbf{F}actor named MOSF.
For the $D$-dimensional scaling factors $\bm{s}$, Figure\,\ref{fig:overview}(d) illustrates the histogram. 
To characterize this set of scaling factors, we can use statistics such as the maximum, minimum, median and mean.
Based on theoretical analysis and experimental verification, we have confirmed that abnormal channels in activations and outliers in scaling factors affect the performance of the model, limiting its overall expressive power.
Our goal is to pick out a statistic that can best represent the majority of the scaling factors while overcoming the influence of a small number of outliers.
Mean absolute deviation (MAD) is a statistical measure of the deviation of data points from a central location. 
Compared to mean squared error (MSE), MAD offers greater robustness and is less sensitive to outliers.
Based on Equation\,(\ref{eq:raparm_13}), We use MAD as a metric to evaluate different statistics. 
\begin{align}
\label{eq:raparm_13}
\text{MAD}={\frac{1}{D}}\sum_{i=1}^{D}\lvert\bm{s}_i-\tilde{s}\rvert  .
\end{align}
Table\,\ref{tab:MAD_Value} presents the MAD values obtained when selecting the optimal scaling factor for layer-wise quantization through the mean or median. 
The experiment shows that the scaling factor $\tilde{s}$ determined by the median results in the smallest mean absolute deviation, and the model performs better.
Therefore, MOSF finds the optimal scaling factor required for layer-wise quantization by using the median.
The optimal $\tilde{s}$ determined by MOSF enables precise quantization for the majority of channels, thus compensating for losses in a few abnormal channels.
The zero point is the same.
%
\begin{align}
\label{eq:raparm_14}
    \tilde{s} = \text{Median}(\bm{s}), ~~\tilde{z} = \text{Median}(\bm{z}) .
\end{align}
\textbf{Superiority.} 
MOSF addresses the accuracy degradation in the transition from channel-wise to layer-wise quantization without adding any extra overhead.
In low-bit settings, MOSF is more effective and does not require parameter optimization.
It can seamlessly integrate into the scale reparameterization framework, providing thorough technical support for quantizing post-LayerNorm activations.

\section{Experimentation}
\label{sec:Experimentation}
\subsection{Experimental Setting}
\label{sec:Experimental Setting}
\textbf{Models and Datasets}. 
To validate the effectiveness of DopQ-ViT, we evaluate the image classification performance of ViT~\cite{Vit}, DeiT~\cite{deit}, and Swin-T~\cite{liu2021swin} models on the ImageNet~\cite{russakovsky2015imagenet}.
Besides, we evaluate the performance of object detection and instance segmentation on the COCO~\cite{lin2014microsoft}, using Mask R-CNN~\cite{he2017mask} and Cascade Mask R-CNN~\cite{cai2018cascade} with Swin-T as the backbone, respectively.

\textbf{Implementation Details.} 
The pretrained models are obtained from the Timm library.
We employ the uniform quantizer for all weights and activations, except post-Softmax activations, which use the Tan quantizer.
We adopt the same setup as the I\&S-ViT~\cite{zhongyunshan}, including block-wise reconstruction, to ensure reliable experimental results.
The calibration dataset includes 1024 randomly selected images.
The optimization uses the Adam optimizer~\cite{kingma2014adam} with an initial learning rate and adjusts the rate using cosine annealing.
For classification experiments on the ImageNet dataset, we set the batch size to 64 and 1000 optimization iterations for all cases, except for the 6-bit case, which is set to 200 iterations.
For detection experiments on the COCO dataset, we only optimize the backbone, set the batch size to 1, and adopt 1000 optimization iterations.
All experiments were conducted on the A6000 GPU.
\begin{table*}[!t]
\vskip -2mm
\caption{Quantization results for image classification on the ImageNet dataset. "Opti." represents the optimization-based approach. "W/A" denotes the quantization bit-width of weights and activations as W bits and A bits, respectively. "3*" indicates quantizing post-Softmax activations to 2-bit.}
\label{tab:imagenet}
\begin{center}
\begin{small}
\begin{tabular}{cccccccccc}
\toprule
\textbf{Method}  & \textbf{Opti.} &\textbf{W/A} & \textbf{ViT-S }& \textbf{ViT-B} & \textbf{DeiT-T }& \textbf{DeiT-S} & \textbf{DeiT-B} & \textbf{Swin-S} & \textbf{Swin-B}\\
\midrule
Full-Precision & - & 32/32 & 81.39 & 84.54 & 72.21 & 79.85 & 81.80 & 83.23 & 85.27\\
\midrule
I\&S-ViT~\cite{zhongyunshan} & $\checkmark$ & 2/3 &17.01 &	38.46 &	12.89 &	25.74 &	59.04 &	60.48 & 	42.88 \\
DopQ-ViT (ours) & $\checkmark$ & 2/3 & 	\textbf{18.98} &	\textbf{41.10} &	\textbf{14.87} &	\textbf{27.02} &	\textbf{61.28} &	\textbf{61.61} &	\textbf{45.46}\\ 
\midrule
I\&S-ViT~\cite{zhongyunshan} &$\checkmark$ &  ~~3/3* &37.88 	&61.93 &	27.28 	& 54.44 	& 71.24 	&72.29 	&65.84 \\
DopQ-ViT (ours) & $\checkmark$ & ~~3/3* &\textbf{41.44}  	& 58.96 &	\textbf{31.34} &	\textbf{54.74} &	\textbf{71.71} &	\textbf{72.89}& 	\textbf{68.26} \\
\midrule
BRECQ~\cite{li2021brecq} & $\checkmark$ & 3/3 & 0.42 & 0.59 & 25.52 & 14.63 & 46.29 & 11.67 & 1.70 \\
QDrop~\cite{wei2021qdrop} & $\checkmark$ & 3/3 & 4.44 & 8.00 & 30.73 & 22.67 & 24.37 & 60.89 & 54.76\\
PD-Quant~\cite{Pd-quant} & $\checkmark$ & 3/3 & 1.77 & 13.09 & 39.97 & 29.33 & 0.94 & 69.67 & 64.32\\
RepQ-ViT~\cite{Repq-vit} & $\times$ & 3/3 & 0.43 & 0.14 & 0.97 & 4.37 & 4.84 & 8.84 & 1.34\\
I\&S-ViT~\cite{zhongyunshan} & $\checkmark$ & 3/3 & 45.16 & 63.77 & 41.52 & 55.78 & 73.30 & 74.20 & 69.30\\
DopQ-ViT (ours) & $\checkmark$ & 3/3 & \textbf{54.72} & \textbf{65.76} & \textbf{44.71} & \textbf{59.26} & \textbf{74.91} & \textbf{74.77} & \textbf{69.63} \\
\midrule
BRECQ~\cite{li2021brecq} & $\checkmark$ & 4/4 & 12.36 & 9.68  & 55.63  & 63.73 & 72.31 & 72.74 & 58.24\\
QDrop~\cite{wei2021qdrop} & $\checkmark$ & 4/4 & 21.24 & 47.30 & 61.93 & 68.27 & 72.60 & 79.58 & 80.93\\
PD-Quant~\cite{Pd-quant}   & $\checkmark$ & 4/4 & 1.51 & 32.45 & 62.46 & 71.21 & 73.76 & 79.87 & 81.12\\
RepQ-ViT~\cite{Repq-vit} & $\times$ & 4/4 & 65.05 & 68.48 & 57.43 & 69.03 & 75.61 & 79.45 & 78.32\\
I\&S-ViT~\cite{zhongyunshan} & $\checkmark$ & 4/4 & 74.87 & 80.07 & 65.21 & 75.81 & 79.97 & 81.17 & 82.60\\
DopQ-ViT (ours) & $\checkmark$ & 4/4 & \textbf{75.69} & \textbf{80.95} & \textbf{65.54} & \textbf{75.84} & \textbf{80.13} & \textbf{81.71} & \textbf{83.34} \\
\midrule
BRECQ~\cite{li2021brecq} & $\checkmark$ & 6/6 & 54.51 & 68.33 & 70.28 & 78.46 & 80.85 &82.02 &83.94\\
QDrop~\cite{wei2021qdrop} & $\checkmark$ & 6/6 & 70.25 & 75.76 & 70.64 & 77.95 & 80.87  &82.60 &84.33\\
PD-Quant~\cite{Pd-quant} & $\checkmark$ & 6/6 & 70.84 & 75.82 & 70.49 & 78.40 & 80.52  &82.51 &84.32\\
RepQ-ViT~\cite{Repq-vit}  & $\times$ & 6/6 & 80.43 & 83.62 & 70.76 & 78.90 & 81.27 &82.79 &84.57\\
I\&S-ViT~\cite{zhongyunshan} & $\checkmark$ & 6/6 & 80.43 & 83.82 & 70.85 & 79.15 & 81.68 & 82.89 & 84.94\\
DopQ-ViT (ours) & $\checkmark$ & 6/6 & \textbf{80.52} & \textbf{84.02} & \textbf{71.17} & \textbf{79.30} & \textbf{81.69} & \textbf{82.95} & \textbf{84.97} \\
\bottomrule
\end{tabular}
\end{small}
\end{center}
\vskip -6mm
\end{table*}
\begin{table*}[!t]
\vskip -2mm
\caption{Quantization results on the COCO dataset. "AP$^\text{box}$" represents the box average precision for object detection, and "AP$^\text{mask}$" represents the mask average precision for instance segmentation. And "*" denotes that the results are obtained from the official code.}
\label{tab:coco}
\begin{center}
\begin{small}
\begin{tabular}{ccccccccccc}
\toprule[1.25pt]
\multirow{3.5}{*}{\textbf{Method}}  & \multirow{3.5}{*}{\textbf{W/A}} & \multicolumn{4}{c}{\textbf{Mask R-CNN}} & \multicolumn{4}{c}{\textbf{Cascade Mask R-CNN}} \\
\cmidrule(lr){3-6}\cmidrule(lr){7-10}
&& \multicolumn{2}{c}{\textbf{w. Swin-T}} & \multicolumn{2}{c}{\textbf{w. Swin-S}} & \multicolumn{2}{c}{\textbf{w. Swin-T}} & \multicolumn{2}{c}{\textbf{w. Swin-S}} \\
&& AP$^\text{box}$ & AP$^\text{mask}$ & AP$^\text{box}$ & AP$^\text{mask}$ & AP$^\text{box}$ & AP$^\text{mask}$ & AP$^\text{box}$ & AP$^\text{mask}$ \\
\midrule[0.75pt]
Full-Precision &   32/32 & 46.0 & 41.6 & 48.5 & 43.3 & 50.4 & 43.7 & 51.9 & 45.0 \\
\midrule[0.1pt]
RepQ-ViT~\cite{Repq-vit} &  3/4 &  22.2  &  24.1  &  27.9  &  29.9  &  40.2  &  35.7  & 43.7 & 38.8\\
DopQ-ViT (Ours) & 3/4 &\textbf{25.4} & \textbf{27.1}  &\textbf{31.0}   &\textbf{32.7}  & \textbf{45.7} & \textbf{39.9}  &\textbf{48.0}  &  \textbf{42.2}  \\ 
\midrule[0.01pt]
BRECQ~\cite{li2021brecq} &   4/4 & 25.4 & 27.6 & 34.9 & 35.4 & 41.2  &  37.0 & 44.5 & 39.2 \\
QDrop~\cite{wei2021qdrop} &   4/4 & 12.4 & 12.9 & 42.7 & 40.2 & 23.9  & 21.2 & 24.1 & 21.4  \\
PD-Quant~\cite{Pd-quant} &   4/4 & 17.7 & 18.1 & 32.2  & 30.9 & 35.5  & 31.0 &  41.6 & 36.3 \\
RepQ-ViT~\cite{Repq-vit} &  4/4 & 36.1 & 36.0 & ~~42.7$^{*}$ & ~~40.1$^{*}$ & 47.0 & 41.4 & 49.3 & 43.1 \\
I\&S-ViT~\cite{zhongyunshan}  & 4/4 &  37.5 & \textbf{36.6} & 43.4 &  40.3 & 48.2 & 42.0& 50.3 & 43.6\\
DopQ-ViT (Ours) & 4/4 &  \textbf{37.5} & 36.5 & \textbf{43.5} &  \textbf{40.4} & \textbf{48.2} & \textbf{42.1} & \textbf{50.3} & \textbf{43.7}\\
\bottomrule[1.0pt]
\end{tabular}
\end{small}
\end{center}
\vskip -6mm
\end{table*}
\subsection{Results on ImageNet Dataset}
We evaluated the performance of DopQ-ViT on the ImageNet for image classification and compared it with other methods, as shown in Table\,\ref{tab:imagenet}. 
\textbf{(i)}
In the 3-bit case, RepQ-ViT fails across all ViT variants.
Optimization-based methods, such as BRECQ, QDrop, and PD-Quant, put in more effort but only enhance the performance of certain ViT variants. 
%
I\&S-ViT has achieved practical performance in 3-bit quantization, but there is still a gap between it and full-precision models.
DopQ-ViT not only improves the accuracy of the 3-bit quantization but also demonstrates stability across various variants.
Specifically, on ViT-S and DeiT-S, DopQ-ViT achieves Top-1 accuracy of 54.72\% and 59.26\%, respectively, with an improvement of 9.56\% and 3.48\% compared to I\&S-ViT.
For DeiT-T, ViT-B, and DeiT-B, DopQ-ViT achieves accuracy of 44.71\%, 65.76\%, and 74.91\%, respectively.
%
%
\textbf{(ii)}
In the 4-bit case, RepQ-ViT achieves more effective quantization.
I\&S-ViT improves efficiency and narrows the gap with full-precision models using reconstruction. 
It is worth noting that DopQ-ViT further enhances the performance and practicality of 4-bit quantized models.
Specifically, it achieves Top-1 accuracy of 75.69\% and 80.95\% for ViT-S and ViT-B, and 65.54\%, 75.84\%, and 80.13\% for DeiT-T, DeiT-S, and DeiT-B, respectively.
\textbf{(iii)}
In the 6-bit case, DopQ-ViT achieves nearly lossless performance in quantization. 
Compared to the full-precision models, the Top-1 accuracy of DeiT-B decreases by only 0.11\%. 
\textbf{(iv)}
We validate the performance of DopQ-ViT under lower bit conditions in W2A3 and W3A3*(only post-Softmax activations are quantized to 2-bit) settings.
Overall, DopQ-ViT achieves outstanding performance across different ViT variants and bit settings. 

\subsection{Results on COCO Dataset}
Object detection and instance segmentation experiments were conducted on the COCO dataset, with the results shown in Table\,\ref{tab:coco}.
The model architecture for these tasks is more complex, which makes quantization challenging.
We conduct experiments in the W4A4 quantization setting.
Specifically, with Swin-T as the backbone of Mask R-CNN, the box AP and mask AP are 37.5\% and 36.5\%, respectively. With Swin-S as the backbone, the box AP and mask AP are 43.5\% and 40.4\%, respectively.
Similarly, with Swin-T as the backbone of Cascade Mask R-CNN, the box AP and mask AP are 48.2\% and 42.1\%, respectively. With Swin-S as the backbone, the box AP and mask AP are 50.3\% and 43.7\%, respectively.
And DopQ-ViT also performs well in W3A4 quantization setting.
%

%
\subsection{Ablation Studies}
\label{subsec: Ablation Studies}
\begin{wraptable}[8]{r}{8cm}
\vspace{-6mm}
\centering
\small
\caption{Ablation studies of different quantizers for post-Softmax activations in 3-bit quantization of DeiT-S model.}
\begin{adjustbox}{width=\linewidth}
\begin{tabular}{cccccc}
\toprule
\multirow{2}{*}{\textbf{Model}}  & \multirow{2}{*}{\textbf{Bit.(W/A)}} & \multicolumn{4}{c}{\textbf{Method}}\\
\cmidrule(lr){3-6}
&& \multicolumn{1}{c}{\textbf{Uniform}} & \multicolumn{1}{c}{\textbf{LogQ}} & \multicolumn{1}{c}{\textbf{SULQ}} & \multicolumn{1}{c}{\textbf{TanQ}}  \\ 
\midrule
\multirow{2}*{ DeiT-S} &  W32/A32   & 79.85  & 79.85 & 79.85  &  79.85  \\ 
\cmidrule(r){2-6}
 &  W32/A3   & 0.38   & 19.14  &  50.94  &   \textbf{55.36}\\ 
\bottomrule
\end{tabular}
\label{tab:Ablation studies of different quantizers}
\end{adjustbox}
\end{wraptable}
\textbf{Effect of TanQ for post-Softmax activations.}   
We use the uniform quantizer to quantize the activations of DeiT-S to 3-bit, with only the post-Softmax activations using the separate quantizer.
Experimental results are shown in Table\,\ref{tab:Ablation studies of different quantizers}.
It's evident that the uniform quantizer is completely ineffective.
%
LogQ can adapt to the distribution of post-Softmax activations.
SULQ further enhances the efficiency.
%
TanQ achieves the accuracy of 55.36\%, which further improves the performance of the low-bit quantized models.

\begin{wraptable}[10]{r}{8cm}
\vspace{-7mm}
\centering
\small
\caption{Ablation studies of Traditional Determination of Scaling Factor (TDSF) and MAD-guided Optimal Scaling Factor (MOSF) in 3-bit quantization for post-LayerNorm activations across different models.}
\begin{adjustbox}{width=\linewidth}
\begin{tabular}{ccc}
\toprule
\multirow{2.5}{*}{\textbf{Quantization}}  & \multirow{2.5}{*}{\textbf{Method}} & \multirow{1}{*}{\textbf{Model}}\\
\cmidrule(lr){3-3}
&& ~\textbf{ViT-S} ~~ \textbf{ViT-B} ~ \textbf{DeiT-T} ~ \textbf{DeiT-S} \\ 
\midrule
Channel-wise   & W32/A3  & 55.70 ~~~ 68.51 ~~~ 46.49 ~~~ 67.66  \\ 
\cmidrule(r){2-3}
\multirow{-2}*{$\downarrow$Reparam} & w. TDSF   & 40.88 ~~~ 61.66 ~~~ 42.76 ~~~ 64.06  \\ 
\multirow{-2}*{Layer-wise} & w. MOSF & \textbf{55.71} ~~~ \textbf{68.52} ~~~ \textbf{46.65} ~~~ \textbf{67.79}   \\ 
\bottomrule
\end{tabular}
\label{tab:TDSF-MOSF}
\end{adjustbox}
\end{wraptable}

\textbf{Effect of MOSF for post-LayerNorm activations.}   
We validate the effectiveness of MOSF through 3-bit quantization experiments across various models, with detailed results shown in Table\,\ref{tab:TDSF-MOSF}. 
The "TDSF" represents the traditional method for determining a single scaling factor for layer-wise quantization.
Compared to TDSF, MOSF effectively mitigates accuracy degradation caused by reparameterization.
With the help of MOSF, each model achieves nearly lossless results after reparameterizing from channel-wise quantization to layer-wise quantization.
%
%
Compared with learning-based methods such as Omniquant~\cite{shao2023omniquant}, MOSF does not require optimization to update parameters.

\textbf{Hyperparameters Analysis.} 
We conduct additional experiments and analysis on parameters $a$ and $b$. 
To avoid redundancy, Table\,\ref{tab:Parameter} only shows the results for each $b$ value with the optimal $a$ value. Despite variations in $b$, the effective coordination between $a$ and $b$ allows the TanQ to perform well, contributing to the model's performance. 
$a$ and $b$ are pre-computed using a calibration dataset, incurring no online cost. 
We perform two rounds of search, with the time costs shown in Table\,\ref{tab:Overhead of Parameter Search}.
The cost is in the hundreds of milliseconds, which is acceptable considering the performance gains.

\begin{wraptable}[7]{r}{8cm}
\vspace{-6mm}
\centering
\small
\caption{Ablation studies on the effectiveness of the Tan Quantizer (TanQ) and MAD-guided Optimal Scaling Factor (MOSF).}
\begin{adjustbox}{width=\linewidth}
\begin{tabular}{cccccc}
\toprule
\multirow{2.5}{*}{\textbf{Model}}  & \multirow{2.5}{*}{\textbf{Bit.(W/A)}} & \multicolumn{4}{c}{\textbf{Method ~~~(TanQ,~MOSF)}}\\
\cmidrule(lr){3-6}
&& \multicolumn{1}{c}{\textbf{($\times$,~$\times$)}} & \multicolumn{1}{c}{\textbf{($\checkmark$,~$\times$)}} & \multicolumn{1}{c}{\textbf{($\times$,~$\checkmark$)}} & \multicolumn{1}{c}{\textbf{($\checkmark$,~$\checkmark$)}}  \\ 
\midrule
Swin-B &  W4/A4   & 67.50    &  71.59 &  	81.30 	  &   \textbf{81.58}\\ 
\bottomrule
\end{tabular}
\label{tab:TanQ and MOSF}
\end{adjustbox}
\end{wraptable}
\textbf{Effect of TanQ and MOSF.} 
We conduct ablation studies on TanQ and MOSF, and the results are shown in Table\,\ref{tab:TanQ and MOSF}.
Based on a baseline accuracy of 67.50\%, TanQ can improve the accuracy to 71.59\%, MOSF can improve it to 81.30\%. With the help of TanQ and MOSF, the final model achieves an accuracy of 81.58\%.
\begin{wraptable}[6]{r}{8cm}
\vspace{-7mm}
\centering
\small
\caption{End to end inference speedup on different models.}
\begin{adjustbox}{width=\linewidth}
\begin{tabular}{ccccc}
\toprule
\textbf{Model}  	 & \textbf{ViT-B} 	 & \textbf{ViT-L} 	 	 & \textbf{DeiT-B} 	 & \textbf{Swin-B} \\
\midrule
FP16 Latency(ms) 		&568 	&1663 		&614 	&724 \\
W4A4 Latency(ms) 	 	&306 	&629 	 	&295	&441\\
SpeedUp 	 	&1.85x 	&2.64x 	 	&2.08x 	&1.64x \\
\bottomrule
\end{tabular}
\label{tab:Hardware Efficiency Comparison}
\end{adjustbox}
\end{wraptable}
\textbf{SpeedUp and Time Efficiency.} 
We modify CUDA kernels from DUQ~\cite{zhong2025towards} for quantization, matrix multiplication, and dequantization, measuring end-to-end inference time. 
Computations for LayerNorm and Softmax are performed at high precision. 
Evaluations are conducted on 200 images using an A6000 GPU, and the results are shown in the Table~\ref{tab:Hardware Efficiency Comparison}.
The latency shows that DopQ-ViT achieves actual speedup, demonstrating its practical applicability for real-world scenarios.
Besides, Figure\,\ref{fig:runtime_and_accuracy} illustrates the runtime and accuracy of PTQ methods. 
DopQ-ViT performs better in time efficiency compared to optimization-based PTQ methods and demonstrates superior accuracy compared to other direct PTQ methods.

\section{Related Works}
\label{Related Works}
\textbf{Vision Transformers (ViTs).} Based on the self-attention mechanism, ViTs have achieved significant success in visual tasks such as image classification~\cite{bhojanapalli2021understanding, chen2021crossvit}, object detection~\cite{li2022exploring, carion2020end}, and instance segmentation~\cite{wang2021end, chen2021simple}. Besides, they also inspired other works~\cite{jiangimage, lin2025toklip, zhou2024doge}.
Compared to Convolutional Neural Networks, ViTs can capture global information from the entire image more effectively and serve as superior vision backbones.
ViT~\cite{Vit} divides images into patches, treats them as sequential data, and employs the self-attention to analyze dependencies within the images.
DeiT~\cite{deit} combines knowledge distillation with ViT to reduce data dependency and achieve superior results.
Swin Transformer~\cite{liu2021swin} constructs a hierarchical vision transformer using shifted windows to efficiently handle features at different scales.

\textbf{Quantization.} The extensive computation and high latency of ViTs limit their applications, while model quantization effectively reduces memory usage and accelerates inference by converting floating-point parameters to integers.
Unlike Quantization-Aware Training (QAT)~\cite{nagel2022overcoming, esser2019learned, gong2019differentiable, li2023vit, liu2023oscillation, li2022q, li20223q, zhong2022dynamic, zhong2023multiquant}, Post Training Quantization (PTQ)~\cite{li2022patch, ding2022towards, frumkin2023jumping, li2023psaq, lin2022fqvit, zhongyunshan, Repq-vit, yuan2022ptq4vit, liu2021post, zhong2024erq, lin2024duquant} can achieve model compression without retraining the network and has received widespread attention.
%
%
PTQ4ViT~\cite{yuan2022ptq4vit} observes the power-law distribution of post-Softmax activations, proposing the twin uniform quantization method to reduce errors.
APQ-ViT~\cite{ding2022towards} proposes the Matthew-effect Preserving Quantization to preserve the Matthew Effect. 
There are also PTQ4ViT~\cite{yuan2022ptq4vit}, FQ-ViT~\cite{lin2022fqvit}, RepQ-ViT~\cite{Repq-vit}, and I\&S-ViT~\cite{zhongyunshan} mentioned above. These PTQ methods all propose customized quantization paradigms.
However, there are still difficulties in maintaining the accuracy of the model under low-bit quantization.
Luckily, the optimization-based PTQ methods can mitigate the accuracy degradation.
Inspired by the works of BRECQ~\cite{li2021brecq}, QDrop~\cite{wei2021qdrop}, PD-Quant~\cite{Pd-quant}, MGRQ~\cite{yang2024mgrq}, we employ block-wise reconstruction to establish a quantization foundation.
%
By comparing the differences between the outputs of the full-precision transformer block and the quantized transformer block, we can optimize the quantization parameters such as scaling factor and zero-point.

\section{Conclusion}
\label{sec:Conclusion}
In this paper we analyze two reasons for the poor performance of post-training quantization methods for ViTs.
%
%
To address these issues, we propose a distribution-friendly and outlier-aware PTQ method named DopQ-ViT.
%
We find that the existing quantizers cannot fit the power-law distribution of post-Softmax activations and ignore values close to 1.
So we propose the Tan Quantizer to focus more on values close to 1 and fit the distribution of post-Softmax activations. 
We also perceive that model performance decreases during the transition from channel-wise to layer-wise quantization of post-LayerNorm activations. 
This is influenced by abnormal channels in the activations and outliers in the scaling factors.
Therefore, we propose MOSF to select the MAD-guided Optimal Scaling Factor, which mitigates the impact of a few abnormal channels by accurately quantizing the majority of channels.
We evaluate DopQ-ViT across different ViT variants and bit settings in classification and detection tasks.
The experimental results indicate that DopQ-ViT has a competitive advantage over other PTQ methods and further enhances the practicality of low-bit quantized models.

\bibliography{reference}
\bibliographystyle{plainnat}
\newpage
\appendix
\section*{Supplementary Materials.}
\begin{figure}[h]
\vskip 0mm
\begin{center}
\centerline{\includegraphics[width=0.8\columnwidth]{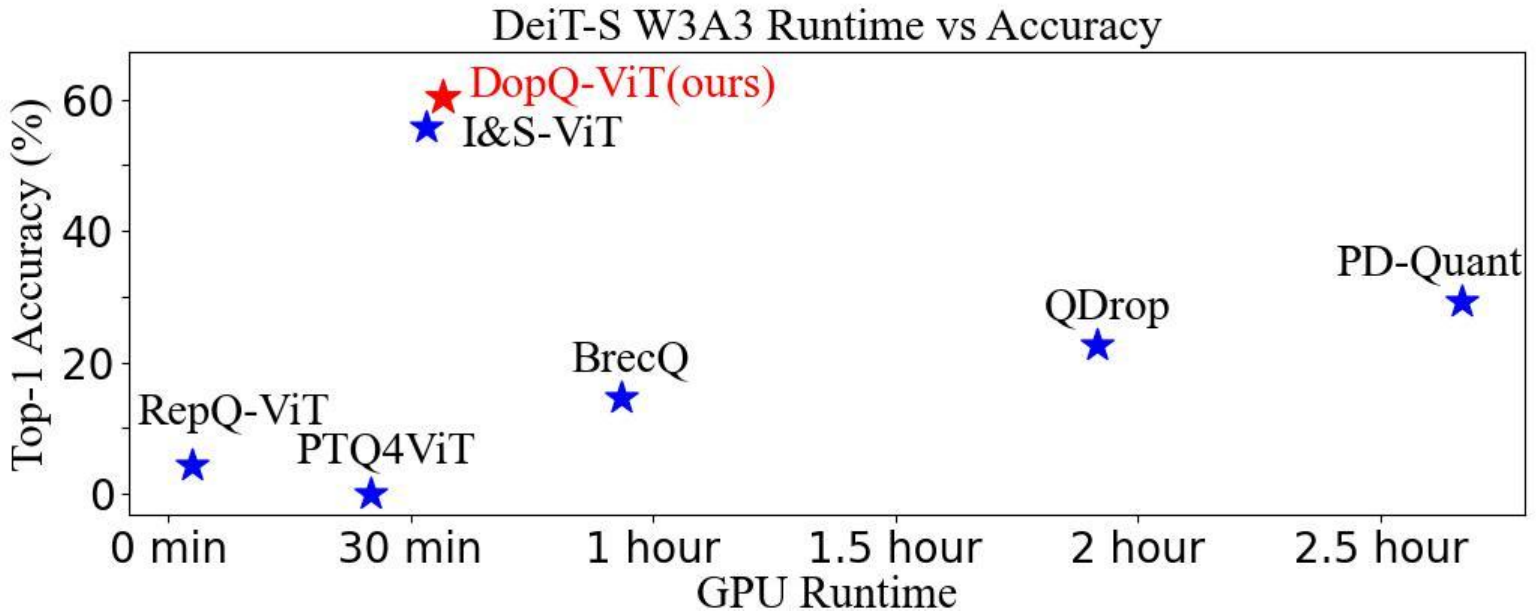}}
\caption{The runtime vs. accuracy of PTQ methods.}
\vskip -2mm
\label{fig:runtime_and_accuracy}
\end{center}
\vskip 4mm
\end{figure}

\begin{figure}[!h]
\begin{center}
\centerline{\includegraphics[width=0.7\columnwidth]{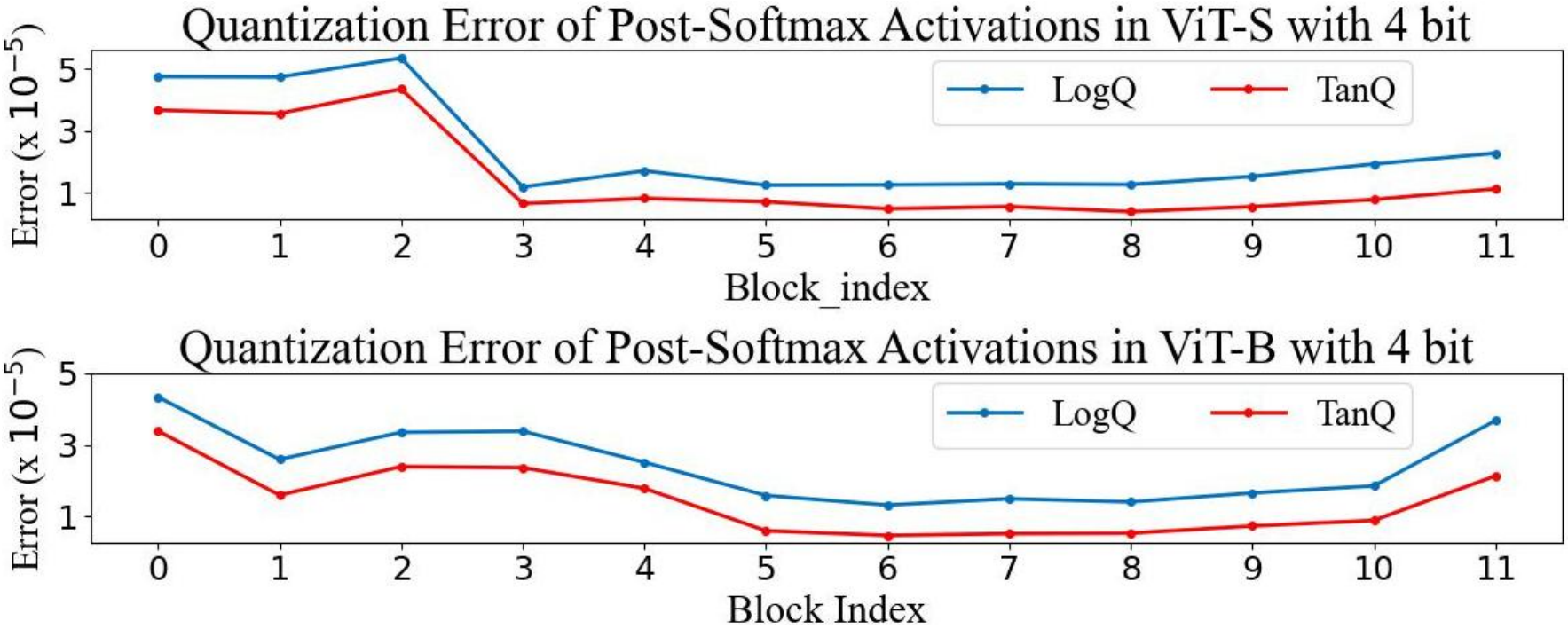}}
\caption{Comparison of LogQ and TanQ for 4-bit quantization errors in post-Softmax activations.}
\label{fig:quantization error}
\end{center}
\vskip 2mm
\end{figure}
\begin{table*}[!h]
\caption{Performance of different W3A3 quantized models under different parameter $b$ values.}
\label{tab:Parameter}
\vskip 0.15in
\begin{center}
\begin{small}
\begin{tabular}{c|cccccccccc}
\toprule
\textbf{Model} & \multicolumn{10}{c}{\textbf{Top-1 Accuracy (\%)} }\\
\midrule
Parameter $b$&0.71  &0.72 &0.73 &0.74 &0.75 &0.76 &0.77 &0.78 &0.79 & 0.80\\
\midrule
DeiT-Tiny &44.55  &43.46  &45.07  &43.71  &44.29  &45.00  &46.02  &45.68  &44.81 &41.20   \\
DeiT-Small&58.30  &58.54  &58.57  &59.11  &57.56  &57.95  &58.99  &58.93  &58.78 &57.88 \\
ViT-Base &64.24  &64.92  &65.16  &64.60  &64.64  &64.66  &64.58  &65.05  &65.38   &64.52\\
\midrule
Parameter $b$ &0.81  &0.82 &0.83 &0.84 &0.85 &0.86 &0.87 &0.88 &0.89 & 0.90 \\
\midrule
DeiT-Tiny  &44.97  &44.98  &45.71 & 44.35& 45.86&46.12 & 45.45&45.64 & 40.10& 45.96 \\
DeiT-Small  &59.05  &58.68  &58.93 & 58.54& 58.88&58.55 &59.03 &59.76 & 54.33& 60.01\\
ViT-Base  &65.03  &65.43  &65.68  &65.14 & 65.00& 65.69& 65.43& 65.59&56.25 &65.70 \\
\midrule
Parameter $b$ &0.91  &0.92 &0.93 &0.94 &0.95 &0.96 &0.97 &0.98 &0.99 & 0.999\\
\midrule
DeiT-Tiny & 45.26& 45.15& 45.23&45.60 &45.11 & 45.25 &43.36 &44.75 &44,71 &44.44\\
DeiT-Small&59.23 & 58.86&59.94 & 59.88&58.51 & 59.91 &59.80 &58.18 &59.26 &58.64\\
ViT-Base  &65.58 &65.03 & 65.22&66.03 &65.02 & 65.49 &65.64 & 65.55 &65.76 &65.62\\
\bottomrule
\end{tabular}
\end{small}
\end{center}
\vskip -0.1in
\end{table*}

\begin{table}[h]
\vskip -2mm
\caption{The time required for parameter search and the performance of model accuracy.}
\label{tab:Overhead of Parameter Search}
\vskip -8mm
\begin{center}
\begin{small}
\vskip -4mm
\begin{tabular}{ccccccccc}
\toprule
\multirow{3}*{\textbf{Search Rounds}} 	& \multicolumn{2}{c}{\textbf{ViT-S}} 	& \multicolumn{2}{c}{\textbf{ViT-B}} 	& \multicolumn{2}{c}{\textbf{DeiT-T}} 	& \multicolumn{2}{c}{\textbf{DeiT-S}} \\
\cmidrule(lr){2-3}\cmidrule(lr){4-5}\cmidrule(lr){6-7}\cmidrule(lr){8-9}
&Accuracy & Time&Accuracy & Time&Accuracy & Time&Accuracy & Time  \\
\midrule
1x Rounds 	&35.6 &85ms 	&52.5 &169ms 	&29.3 &45ms 	&52.8 &85ms \\
2x Rounds 	&51.3 &175ms 	&66.2 &346ms 	&45.1 &91ms 	&60.1 &175ms 	\\
\bottomrule
\end{tabular}
\end{small}
\end{center}
\end{table}
\begin{table*}[!h]
\caption{Comparison of the MAD for different methods of post-LayerNorm activations.}
\label{tab:MAD_Value}
\begin{center}
\begin{small}
\begin{tabular}{c|cc|cccccc}
\toprule
{\textbf{Model}}  &{\textbf{Structure}}  &{\textbf{Method}}& \multicolumn{6}{c}{\textbf{3bit Quantization MAD$\downarrow$}}\\
\midrule
\multirow{10}*{\textbf{ViT-S}}  &Block Index & -& \multicolumn{1}{c}{$0_{th}$}& \multicolumn{1}{c}{$1_{th}$}  & \multicolumn{1}{c}{$2_{th}$} & \multicolumn{1}{c}{$3_{th}$} & \multicolumn{1}{c}{$4_{th}$} & \multicolumn{1}{c}{$5_{th}$} \\
&\multirow{2}*{{LayerNorm1}}  &Mean & 0.135 	 &	0.084 	 &	0.077 	 &	 	0.065  &		0.084 	  &		0.085\\
&  &Median   & \textbf{0.134}  	 	&\textbf{0.081}  	 	& \textbf{0.069}  	 &	\textbf{0.063}  	& \textbf{0.075}  	& 	\textbf{0.077}  \\
&\multirow{2}*{{LayerNorm2}}  &Mean  & 0.365 	 &		0.083 	 &	0.062  &		0.083 	 &	0.070 	 	 &	0.072 \\
&  &Median  &\textbf{0.363}  	 &	\textbf{0.077}  	&\textbf{0.057}  	& 	\textbf{0.072}  	&\textbf{0.058}  	 &	\textbf{0.060}  	\\
\cmidrule{2-9}
&Block Index & -& \multicolumn{1}{c}{$6_{th}$} & \multicolumn{1}{c}{$7_{th}$} & \multicolumn{1}{c}{$8_{th}$} & \multicolumn{1}{c}{$9_{th}$} & \multicolumn{1}{c}{$10_{th}$} & \multicolumn{1}{c}{$11_{th}$}  \\ 
&\multirow{2}*{{LayerNorm1}}  &Mean   &	0.086 	    &	 0.081 &	 0.075 		 &	0.070 	 &	0.072  &		0.070\\
&  &Median  &\textbf{0.079}  &	 	\textbf{0.076}  	 	&\textbf{0.070}  	 &	\textbf{0.065}  		&\textbf{0.066}  	& 	\textbf{0.068}  	\\
&\multirow{2}*{{LayerNorm2}}  &Mean  &	0.065 	 &	 	0.072 	 &	0.155 	 &	0.214 	 &	0.472 	 &	0.067\\
&  &Median  &\textbf{0.056}  	 	&\textbf{0.064}  	&\textbf{0.139}  	 &	\textbf{0.192}  	&\textbf{0.455}  	& 	\textbf{0.062} \\
\midrule
\multirow{10}*{\textbf{ViT-B}}  &Block Index & -& \multicolumn{1}{c}{$0_{th}$}& \multicolumn{1}{c}{$1_{th}$}  & \multicolumn{1}{c}{$2_{th}$} & \multicolumn{1}{c}{$3_{th}$} & \multicolumn{1}{c}{$4_{th}$} & \multicolumn{1}{c}{$5_{th}$} \\
&\multirow{2}*{{LayerNorm1}}  &Mean  & 	0.290 &	 0.112&	0.134 &	 0.146 & 	0.171 	 &	0.205\\
&  &Median  &\textbf{0.264} &	 \textbf{0.110} &	\textbf{0.132} &	 \textbf{0.140} & 	\textbf{0.167}  &	\textbf{0.203} 	\\
&\multirow{2}*{{LayerNorm2}}  &Mean  & 	0.701  &	0.267 	 &	0.176 	 &	 	0.117 & 	0.133 	 &	0.121 \\
&  &Median  &\textbf{0.598}  &	\textbf{0.261} &	\textbf{0.176} &	 \textbf{0.111} 	& 	\textbf{0.118} &	\textbf{0.112} 	\\
\cmidrule{2-9}
&Block Index & -& \multicolumn{1}{c}{$6_{th}$} & \multicolumn{1}{c}{$7_{th}$} & \multicolumn{1}{c}{$8_{th}$} & \multicolumn{1}{c}{$9_{th}$} & \multicolumn{1}{c}{$10_{th}$} & \multicolumn{1}{c}{$11_{th}$}  \\ 
&\multirow{2}*{{LayerNorm1}}  &Mean  &	0.228  &	0.232 & 	0.192 &	0.160  &	0.139 	 &	 0.139\\
&  &Median  &\textbf{0.227} 	 &	 \textbf{0.232} 	& 	\textbf{0.190} 	 &	\textbf{0.155} &	\textbf{0.132} 	 &	\textbf{0.139} 	\\
&\multirow{2}*{{LayerNorm2}}  &Mean  &	 	0.128 	 &	 0.130 & 	0.247 	 &	0.440 	 &	0.654 	 &	 0.131\\
&  &Median  &\textbf{0.117} 	 &	 \textbf{0.119} 	& 	\textbf{0.232} 	 	 &	\textbf{0.421} 	 &	\textbf{0.645} &	\textbf{0.120}\\
\midrule
\multirow{10}*{\textbf{DeiT-T}}  &Block Index & -& \multicolumn{1}{c}{$0_{th}$}& \multicolumn{1}{c}{$1_{th}$}  & \multicolumn{1}{c}{$2_{th}$} & \multicolumn{1}{c}{$3_{th}$} & \multicolumn{1}{c}{$4_{th}$} & \multicolumn{1}{c}{$5_{th}$} \\
&\multirow{2}*{{LayerNorm1}}  &Mean   &0.185 		&0.125 	&0.155 		&0.111 	&0.108 		&0.096\\
&  &Median  &\textbf{0.184}  	 	&\textbf{0.124} 	 	 &\textbf{0.145}  	 	&\textbf{0.110}  	 &\textbf{0.105}  	 	&\textbf{0.092}  	\\
&\multirow{2}*{{LayerNorm2}}  &Mean  &0.100 	&0.097 	&0.074 	    &0.064 	&0.057 		&0.053 \\
&  &Median  &\textbf{0.096}  	 	&\textbf{0.090}  	&\textbf{0.071}  	 	&\textbf{0.063}  	&\textbf{0.056}  	    &\textbf{0.052}\\
\cmidrule{2-9}
&Block Index & -& \multicolumn{1}{c}{$6_{th}$} & \multicolumn{1}{c}{$7_{th}$} & \multicolumn{1}{c}{$8_{th}$} & \multicolumn{1}{c}{$9_{th}$} & \multicolumn{1}{c}{$10_{th}$} & \multicolumn{1}{c}{$11_{th}$}  \\ 
&\multirow{2}*{{LayerNorm1}}  &Mean  &0.319 	 	&0.092 	&0.072 		&0.115 	&0.129 	 	&0.139 	\\
&  &Median  &\textbf{0.308}  	 	&\textbf{0.092}  	&\textbf{0.072}  	 	&\textbf{0.109} 	&\textbf{0.113} 	 	&\textbf{0.121}  	\\
&\multirow{2}*{{LayerNorm2}}  &Mean  &0.137 	 	&0.145 	&0.157 	 	&0.126 	&0.081 		&0.061 \\
&  &Median  &\textbf{0.119}  	 	&\textbf{0.130}  	&\textbf{0.150}  	 	&\textbf{0.118}  	&\textbf{0.078}  	 	&\textbf{0.059}  \\
\midrule
\multirow{10}*{\textbf{DeiT-S}}  &Block Index & -& \multicolumn{1}{c}{$0_{th}$}& \multicolumn{1}{c}{$1_{th}$}  & \multicolumn{1}{c}{$2_{th}$} & \multicolumn{1}{c}{$3_{th}$} & \multicolumn{1}{c}{$4_{th}$} & \multicolumn{1}{c}{$5_{th}$} \\
&\multirow{2}*{{LayerNorm1}}  &Mean  &0.196 	 	&0.107 	&0.084 	 	&0.075 	&0.077 	 	&0.077 	\\
&  &Median  &\textbf{0.194} 		&\textbf{0.106} 	&\textbf{0.083} 	    &\textbf{0.074} 	&\textbf{0.075} 		&\textbf{0.074} 	\\
&\multirow{2}*{{LayerNorm2}}  &Mean  &0.083 	 	&0.068 	&0.058 	 	&0.055 	&0.047 		&0.041 	\\
&  &Median  &\textbf{0.078}	 	&\textbf{0.066} 	&\textbf{0.055} 		&\textbf{0.054} 	&\textbf{0.045} 	 	&\textbf{0.040} 	\\
\cmidrule{2-9}
&Block Index & -& \multicolumn{1}{c}{$6_{th}$} & \multicolumn{1}{c}{$7_{th}$} & \multicolumn{1}{c}{$8_{th}$} & \multicolumn{1}{c}{$9_{th}$} & \multicolumn{1}{c}{$10_{th}$} & \multicolumn{1}{c}{$11_{th}$}  \\ 
&\multirow{2}*{{LayerNorm1}}  &Mean  &0.453 	&0.122 	&0.086 	&0.082 	&0.098 	&0.116\\
&  &Median  & \textbf{0.450} 	 	&\textbf{0.122} 	&\textbf{0.085} 	 	&\textbf{0.077} 	&\textbf{0.084} 	 	&\textbf{0.096} 	\\
&\multirow{2}*{{LayerNorm2}}  &Mean  &0.128 	&0.194 	&0.178 	&0.126 	&0.079 	&0.045\\
&  &Median  &\textbf{0.111} 	 	&\textbf{0.174} 	&\textbf{0.164} 	 	&\textbf{0.119} 	&\textbf{0.076} 	 	&\textbf{0.043}  \\
\bottomrule
\end{tabular}
\end{small}
\end{center}
\end{table*}

\end{document}